\pdfoutput=1

\documentclass[11pt]{article}
\usepackage[marginal]{footmisc}
\usepackage{amsmath}
\usepackage{xspace}
\usepackage{caption}
\usepackage{float} 
\usepackage{subcaption}
\usepackage[final]{acl}

\usepackage{times}
\usepackage{latexsym}

\usepackage[T1]{fontenc}

\usepackage[utf8]{inputenc}

\usepackage{microtype}

\usepackage{inconsolata}
\usepackage{CJKutf8}
\usepackage{booktabs}

\usepackage{graphicx}

\usepackage{tablefootnote}
\usepackage{hyperref}

\newcommand{\mname}{{\text LayAlign}\xspace}
\newcommand{\aname}{{\text Layer-Wise Adaptive Fusion and Alignment Strategy}\xspace}
\newcommand{\attaname}{{\text adaptive fusion-enhanced attention}\xspace}

\title{LayAlign: Enhancing Multilingual Reasoning in Large Language Models via Layer-Wise Adaptive Fusion and Alignment Strategy}

\author{%
    Zhiwen Ruan$^{1}$\thanks{Equal Contribution.}, \
    Yixia Li$^{1}$\footnotemark[1], \
    He Zhu$^{2}$,\
    Longyue Wang$^{3}$, Weihua Luo$^{3}$ \\
    \textbf{Kaifu Zhang}$^{3}$, \textbf{Yun Chen}$^{4}$, \textbf{Guanhua Chen}$^{1}$\thanks{Corresponding Author.}\\
    $^1$Southern University of Science and Technology,
    $^2$Peking University\;\\
    $^3$Alibaba International Digital Commerce,$^4$Shanghai University of Finance and Ecnonmics\; \\
}

\begin{document}
\maketitle
\begin{CJK*}{UTF8}{gbsn}

\begin{abstract}

Despite being pretrained on multilingual corpora, large language models (LLMs) exhibit suboptimal performance on low-resource languages. Recent approaches have leveraged multilingual encoders alongside LLMs by introducing trainable parameters connecting the two models. However, these methods typically focus on the encoder's output, overlooking valuable information from other layers. 
We propose \aname (\mname), a framework that integrates representations from all encoder layers, coupled with the \attaname mechanism to enable layer-wise interaction between the LLM and the multilingual encoder. Extensive experiments on multilingual reasoning tasks, along with analyses of learned representations, show that our approach consistently outperforms existing baselines.

\end{abstract}

\section{Introduction}
Large Language Models (LLMs) are predominantly trained on corpora emphasizing a select group of high-resource languages, enabling them to demonstrate strong reasoning capabilities in tasks such as mathematics \cite{metamath, llemma, wizardmath, orcamath} and commonsense reasoning \cite{commonsense_1, commonsense_2}. However, most of these LLMs are derived from English-centric models and fine-tuned using English-specific downstream data. Consequently, their performance in low-resource languages remains significantly limited, leading to pronounced disparities between high-resource and low-resource language capabilities.

While multilingual pretrained models attempt to bridge this gap by supporting a broader set of languages, they often exhibit limited reasoning abilities due to constrained training data and model parameters \cite{mT5,nllb}. In contrast, English-centric LLMs display strong reasoning skills but struggle with multilingual understanding, leading to poorer performance in low-resource languages. Inspired by multimodal approaches \cite{flamingo,llava,minigptv,multi-model-1}, works like LangBridge \cite{langbridge} and MindMerger \cite{mindmerger} aim to enhance multilingual reasoning by integrating a multilingual encoder \cite{mT5} with an LLM via a trainable adapter. However, these methods focus only on the top multilingual encoder layer, overlooking the potential richness of intermediate representations.

In this paper, we introduce \aname (\mname), a framework that integrates representations from all multilingual encoder layers by applying distinct fusion ratios for each LLM layer. This approach enables the model to leverage both low- and high-level representations effectively. To incorporate the fused multilingual representations into the decoder-only LLM, we propose a \attaname mechanism combining cross-attention and self-attention. This mechanism uses representations from the layer-wise aligner to generate key-value pairs, with learnable gate parameters modulating cross-attention intensity. 

\mname is optimized with a two-stage finetuning scheme, keeping both the multilingual encoder and LLM backbone frozen. \mname encourages the model to select representations from appropriate encoder layers, facilitating a shared multilingual representation space across all LLM layers.
We evaluate the effectiveness of \mname on mathematical and commonsense reasoning tasks. Our experimental results and analyses of the learned representation space demonstrate that \mname significantly improves reasoning performance for low-resource languages while maintaining strong results for high-resource languages.\footnote{Our code and models are publicly available at \url{https://github.com/sustech-nlp/LayAlign}.}

\section{Related Work}

\subsection{Multilingual Large Language Models}

To address the demand for supporting global linguistic diversity, researchers have expanded into multilingual LLMs \cite{qin2025survey}. Advanced models like Qwen2 \cite{qwen2} and LLaMA3 \cite{llama3} support multiple languages, showcasing robust multilingual capabilities. However, these models are trained from scratch, which incurs substantial computational costs and requires extensive datasets for relevant languages, often leading to inadequate support for low-resource languages. These meticulously trained models frequently face challenges in scaling to other languages, particularly those with lower representation in the training data.

Recently, LangBridge \cite{langbridge} and MindMerger \cite{mindmerger} feature an English-centric LLM backbone, a multilingual encoder that offers multilingual information, and an adapter that facilitates interoperability between the multilingual and English languages. However, these approaches are limited to representations from the topmost encoder layer, neglecting potentially valuable insights from other layers. Our \mname framework follows this line and explores to better leverage the multilingual information of different encoder layers to enhance the multilingual reasoning abilities of LLMs.

\subsection{Aligning Pretrained Representations} 
The integration of encoders with large language models (LLMs) has been widely studied in the cross-modal domain \cite{llava, minigptv, multi-model-1}. Many approaches utilize vision-to-language adapter modules to align visual and textual modalities, mapping the output of vision encoders to the soft prompt inputs of LLMs. Other works employ cross-attention mechanisms to enable more direct interaction between image and text representations \cite{flamingo, evlm}. 
Drawing inspiration from these cross-modal strategies, our method enhances multilingual reasoning by integrating a multilingual encoder with an LLM. To bridge the gap between these components, we introduce an aligner that enables efficient interaction via cross-attention.

\section{Method}

We introduce \mname, which facilitates the direct interaction between all layers of the LLM and the representations of the multilingual encoder through a layer-wise aligner and \attaname. This approach allows for a more comprehensive integration of language comprehension information from the encoder, thereby enhancing the multilingual reasoning capabilities of LLM.
In the subsequent sections, we provide a detailed overview of our framework, focusing on the model architecture (Section~\ref{sec:model_architecture}), \attaname (Section~\ref{sec:gate_attention}), and training methodology (Section~\ref{sec:two_stage_training}).

\subsection{Model Architecture}
\label{sec:model_architecture}

\begin{figure*}[t]
  \includegraphics[width=\linewidth]{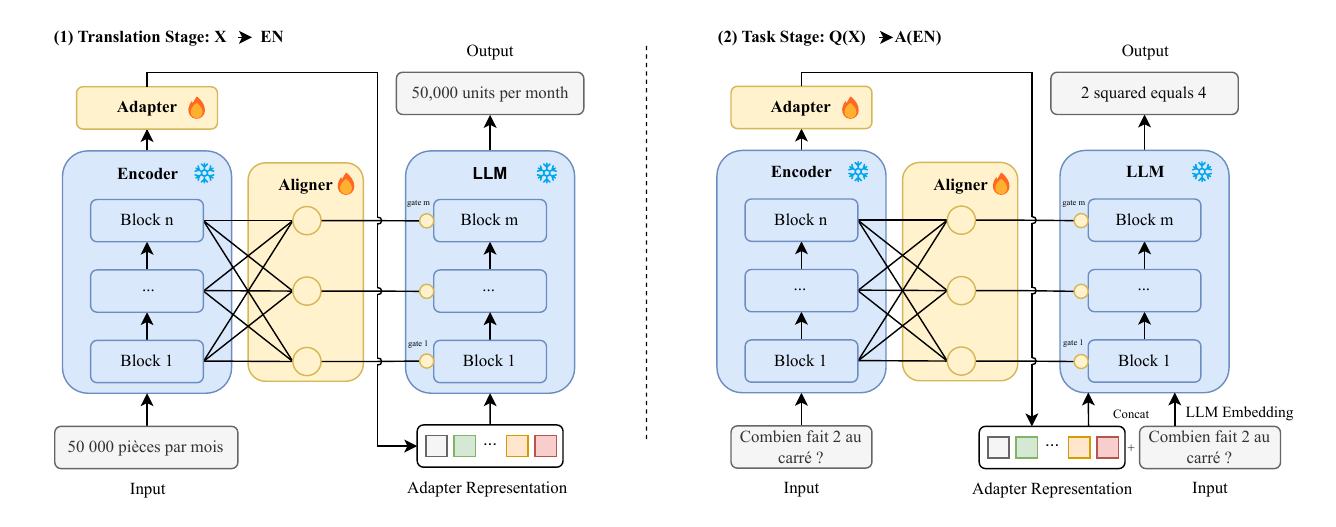}
  \caption{Overview of \mname. A multilingual encoder is aligned with the target LLM with an adapter and the layer-wise aligner. We keep the multilingual encoder and LLM frozen, whereas the adapter and layer-wise aligner are optimized in two stages. For simplicity, shifted output tokens were omitted from the input representation. \textbf{Left:} (1) Translation stage. In this stage, \mname is fine-tuned using translation data, where the data consists of translations from other languages into English. \textbf{Right:} (2) Task Stage. In this stage, \mname is fine-tuned using specialized downstream task data, where the input is multilingual and the output is in English.}
  \label{fig:main_fig}
\end{figure*}

As depicted in Figure~\ref{fig:main_fig}, the adapter and layer-wise aligner are designed to align a multilingual encoder with $n$ layers to the representation space of an LLM with $m$ layers. The input multilingual text $I_\mathrm{in}$ is processed by the encoder, producing a series of representations $\{H_1, H_2, \dots, H_n\}$, where $H_i$ denotes the output of the $i$-th encoder layer.
Following prior work \cite{langbridge}, an adapter is employed to map the final layer's representation $H_n$ to the soft prompt input $I_\mathrm{map}$ for the LLM, thereby enhancing multilingual reasoning capabilities, where $I_\mathrm{map} = \mathrm{Adapter}(H_n)$.

However, this approach only utilizes the final layer representation $H_n$, disregarding the intermediate representations from the embedding $H_0$ through the encoder layers ${H_1, \dots, H_{n-1}}$. To fully harness the multilingual potential of the encoder, we propose a novel layer-wise aligner that explicitly integrates both low-level and high-level representations from multiple layers of the multilingual encoder, rather than relying solely on the final layer's output.

For each LLM layer, the layer-wise aligner generates a fused representation by assigning distinct weights to different multilingual encoder layers. This mechanism allows the model to learn the optimal combination of low-level and high-level features across encoder layers, establishing a correspondence between the multilingual encoder and each LLM layer. While the adapter leverages the final layer representation of the encoder, the layer-wise aligner integrates information from the embedding layer and intermediate encoder layers, enriching the LLM with additional multilingual context. Formally, the fusion process is defined as:
\begin{equation}
  \label{eq:Q_1}
  H_i^K,H_i^V=f_i(H_0,...,H_{n-1}),
\end{equation}
where $f_i\left(\cdot\right)$ is the fusion function for $i$-th layer of LLM, responsible for fusing $\{H_0,...,H_{n-1}\}$ into the fused representations $\{H_i^K,H_i^V\}$. Specifically, $f_i\left(\cdot\right)$ consists of two linear layers with a ReLU activation in between. The resulting $H_i^K$ and $H_i^V$ are then fed into the $i$-th layer of the LLM as keys and values for cross-attention computing. The detailed procedure is provided in Section~\ref{sec:gate_attention}.

\subsection{Adaptive Fusion-Enhanced Attention}
\label{sec:gate_attention}
\begin{figure}[t]
  \includegraphics[width=\linewidth]{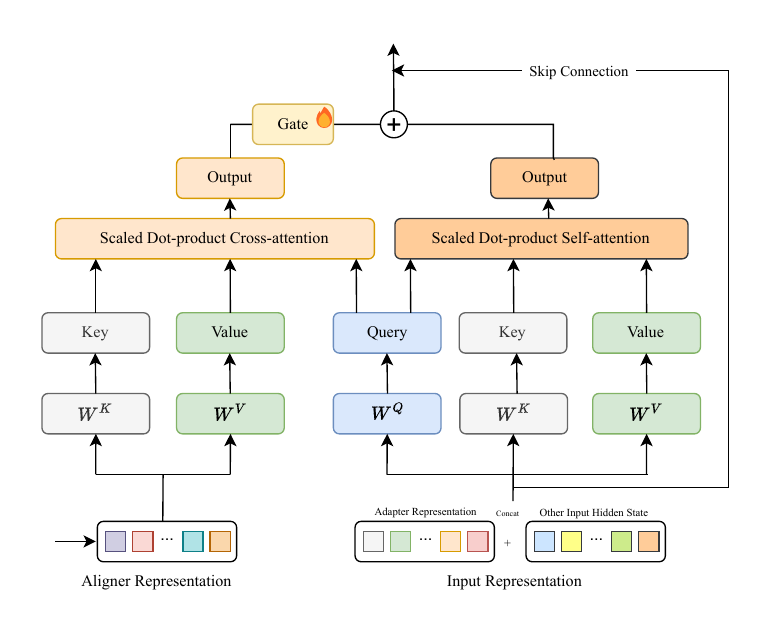}
  \caption{The illustration of our proposed Adaptive Fusion-Enhanced Attention. It consists of self-attention (right), cross-attention (left), and a gate module. Both cross-attention and self-attention modules share the same linear weights as that of the backbone LLM. }
  \label{fig:gate_att}
\end{figure}

We denote the hidden states in the $i$-th LLM decoder layer as $T_i$, where $i \in [0,m]$ and $T_0$ denotes the concatenation of the output from the decoder's embedding layer with the output from the adapter module. The final representation $T_m$ is utilized to generate the next token. In these transformer layers, standard self-attention (SA) is employed.
However, it can not directly interact with the fused representations from the multilingual encoder. To address this limitation, we replace the vanilla attention mechanism in all transformer layers with \attaname, which incorporates self-attention, cross-attention, and a gate module, as shown in Figure~\ref{fig:gate_att}. 

Specially, for the $i$-th LLM layer representation, the attention mechanism is computed as follows:
\begin{align}
    GA(T_{i-1},H_i^K,H_i^V) =SA(T_{i-1})\notag\\
     +g_i \cdot CA(T_{i-1},H_i^K,H_i^V)
\end{align}
Here, keys $H_i^K$ and values $H_i^V$ represent the fused multilingual encoder representations generated by the layer-wise aligner for the $i$-th layer of the LLM. $T_{i-1}$ is the output of the $i-1$-th layer of the LLM, while SA and CA denote self-attention and cross-attention, respectively. 
A learnable gate, $g_i$, is introduced to regulate the incorporation of fused information into the $i$-th layer automatically. This gate is initialized to $0$ to ensure smooth training at the initial stages.
The cross-attention module shares the same linear parameters $W^Q$, $W^K$, $W^V$, and $W^O$ as the self-attention module, thus eliminating the need for additional parameters. These parameters remain frozen during training. 
For clarity, the output linear matrix $W^O$ is omitted in Figure~\ref{fig:gate_att}. 

\subsection{Two-stage Training}
\label{sec:two_stage_training}
The training process is structured into two distinct phases, as illustrated in Figure~\ref{fig:main_fig}. The first phase, referred to as the \textbf{translation stage}, concentrates on aligning the representation spaces between the multilingual encoder and the LLM. During this stage, \mname is fine-tuned using parallel corpora from the many-to-English machine translation task. The input to the LLM is derived from the adapter’s output $I_\mathrm{map}$, denoted as $X=[\langle bos \rangle; I_\mathrm{map}; \langle sep \rangle]$.

The second phase termed the \textbf{task stage}, is designed to enhance the model's performance on specific downstream tasks within a multilingual context. In this phase, \mname is fine-tuned on specialized downstream task data, where the input is multilingual and the output is in English. Here, the LLM's input combines both the adapter’s output and the original user input text $I_\mathrm{in}$ (as shown in Figure~\ref{fig:main_fig}), represented as $X = [\langle bos \rangle; I_\mathrm{map}; \langle sep \rangle; \mathrm{embed}(I_\mathrm{in})]$. Unlike baseline approaches such as LangBridge \cite{langbridge}, which rely solely on the adapter's output $I_\mathrm{map}$ as the LLM input, this approach incorporates additional context, fostering task-specific adaptation and improved multilingual performance. The impact of the additional LLM input $I_\mathrm{in}$ is further examined in Section~\ref{sec:eff_llm_input}.

\section{Experiments}\label{sec:experiments}

We compare LayAlign with baselines on mathematical reasoning, commonsense reasoning, and language understanding tasks following prior studies \cite{mindmerger, langbridge}. 

\subsection{Mathematical Reasoning}

\subsubsection{Experimental Setup}
\label{sec:math_exp}
\paragraph{Evaluation Dataset.} In line with \citet{mindmerger}, we employ two datasets for evaluating \mname: MGSM \cite{weak_multilingual_reasoning_1} and MSVAMP \cite{MathOctopus}. MGSM contains multilingual grade school-level mathematical word problems, while MSVAMP serves as an out-of-domain evaluation set, providing a broader assessment of the multilingual mathematical reasoning capabilities. Models are evaluated using a zero-shot approach.

\paragraph{Training Datasets.} Consistent with the setup in MindMerger \cite{mindmerger}, we leverage the same training data for \mname. In the first stage, the model is trained on translation data from the Lego-MT corpus \cite{lego-mt}, which translates multilingual inputs into English. In the second stage, we employ the composite multilingual mathematical data, referred to as MultilingualMath \cite{metamath, MathOctopus}, consisting of 30,000 samples per language across ten languages. This dataset supports comprehensive training for robust multilingual mathematical reasoning.

\paragraph{Baselines.} We compare our approach against seven baselines.

\begin{table}[]
\resizebox{\columnwidth}{!}{
    \begin{tabular}{llll}
    \toprule
    Method        & Backbone      & Training Data  & Source \\
    \hline
    \multicolumn{4}{l}{\textbf{English-Only Data Baselines}}         \\
    MetaMath      & LLaMA2-7B     & MetaMathQA     & Official checkpoint\tablefootnote{\scriptsize \url{https://huggingface.co/meta-math/MetaMath-7B-V1.0}}   \\
    LangBridge-EN & \texttt{mT5-xl}+MetaMath  & MetaMath-200k  & \citet{langbridge}   \\
    Translate     & NLLB+MetaMath & None           & Reimplementation  \\
    \hline
    \multicolumn{4}{l}{\textbf{Multi-lingual Data Baselines}}        \\
    MetaMath-Mul  & MetaMath      & MultilingualMath   & Reimplementation  \\
    MathOctopus   & LLaMA2-7B     & MGSM8KInstruct & Official checkpoint\tablefootnote{\scriptsize \url{https://huggingface.co/Mathoctopus/Parallel_xRFT_7B}}   \\
    LangBridge    & \texttt{mT5-xl}+MetaMath  & Lego-MT+MultilingualMath   & Reimplementation  \\
    MindMerge     & \texttt{mT5-xl}+MetaMath  & Lego-MT+MultilingualMath   & Official checkpoint\tablefootnote{\scriptsize \url{https://github.com/CONE-MT/MindMerger}}   \\
    \bottomrule
    \end{tabular}    
}
\caption{Comparisions of baselines. LangBridge and MindMerge are trained with the same two-stage data as \mname.}
\vspace{-15pt}
\end{table}

\noindent $\bullet$ \textbf{MetaMath}: \textbf{MetaMath} is fine-tuned from LLaMA2-7B on MetaMathQA, a mathematical dataset derived from GSM8K \cite{gsm8k} and MATH \cite{MATH}. We further train MetaMath on our second phase multilingual task data, resulting in \textbf{MetaMath-Mul}. 

\noindent $\bullet$ \textbf{Translate} \cite{weak_multilingual_reasoning_1}: 
a training-free method that translates the prompt into English for MetaMath. We utilize NLLB-200-3.3B \cite{nllb} as the translator followed MindMerger.

\noindent $\bullet$ \textbf{MathOctopus} \cite{chen2023breakinglanguagebarriersmultilingual}: fine-tuned from LLaMA2-7B on a custom multilingual mathematical reasoning dataset. We utilize their best-performing checkpoint xRFT-MathOctopus.

\noindent $\bullet$ \textbf{LangBridge} \cite{langbridge}: aligns \texttt{mT5-xl} with MetaMath by projecting the final-layer hidden states of \texttt{mT5-xl} into MetaMath’s input via an adapter. We compare against both \textbf{LangBridge-EN}, the original model trained on the English dataset MetaMath-200k, and \textbf{LangBridge}, which we trained on the same datasets as \mname using our two-stage training process for a fair comparison.

\noindent $\bullet$ \textbf{MindMerger} \cite{mindmerger}: it shares a similar architecture with LangBridge. While LangBridge feeds multilingual math prompts exclusively into \texttt{mT5-xl}, MindMerger processes the prompts in parallel through both \texttt{mT5-xl} and MetaMath at the second stage.

\paragraph{Model and Training Details.} 
We utilize the encoder of \texttt{mT5-xl} as the multilingual encoder, comprising 1.6 billion parameters, with MetaMath \cite{metamath} serving as the LLM. The training procedure is conducted in two stages. During the first stage, the learning rate is set to $4 \times 10^{-5}$, with a batch size of 128, over 3 epochs, and a warmup ratio of 0.05. In the second stage, the learning rate is adjusted to $3 \times 10^{-5}$, while maintaining the batch size at 128, the number of epochs at 3, and the warmup ratio at 0.05. All experiments are executed on 8 NVIDIA L40 GPUs, with the first and second stages taking 9 and 8 hours, respectively.

\subsubsection{Results}

\begin{table*}[]
    \resizebox{\textwidth}{!}{
\begin{tabular}{l|ccc|cccccccccc}
\hline
\textbf{MGSM} & \textbf{Avg.} & \textbf{Lrl.} & \textbf{Hrl.} & \textbf{Bn} & \textbf{Th} & \textbf{Sw} & \textbf{Ja} & \textbf{Zh} & \textbf{De} & \textbf{Fr} & \textbf{Ru} & \textbf{Es} & \textbf{En} \\ \hline
MetaMath & 37.9 & {5.9} & {51.6} & 6.4 & 6.4 & 4.8 & 34.8 & 39.2 & 56.4 & 55.6 & 51.6 & 55.2 & \textbf{68.4} \\
LangBridge-EN & 50.2 & {45.5} & {52.3} & 42.8 & 50.4 & 43.2 & 40.0 & 45.2 & 50.8 & 52.4 & 56.4 & 58.0 & 63.2 \\
Translate & 43.1 & {36.1} & {46.1} & 46.4 & 27.2 & 34.8 & 28.4 & 34.8 & 48.8 & 44.0 & 42.4 & 55.6 & 68.4 \\ \hline
MetaMath-Mul & 38.4 & {35.1} & {39.8} & 32.0 & 36.8 & 36.4 & 35.2 & 40.0 & 40.8 & 41.2 & 39.6 & 40.8 & 41.2 \\
MathOctopus & 40.0 & {33.5} & {42.8} & 30.4 & 35.2 & 34.8 & 38.0 & 45.6 & 41.6 & 38.4 & 39.6 & 46.0 & 50.4 \\
LangBridge & 54.0 & {50.1} & {55.6} & 48.8 & 49.2 & 52.4 & 50.0 & 53.6 & 56.0 & 54.0 & 58.0 & 58.0 & 59.6 \\
MindMerger & {57.4} & {54.8} & {58.6} & {51.2} & {56.8} & {56.4} & {50.8} & {54.4} & {60.0} & {55.2} & \textbf{62.4} & {59.6} & {67.6} \\
\textbf{\mname} & \textbf{59.0} & \textbf{56.4} & \textbf{60.2} & \textbf{51.6} & \textbf{59.2} & \textbf{58.4} & \textbf{52.0} & \textbf{56.0} & \textbf{62.0} & \textbf{61.6} & 61.6 & \textbf{61.6} & 66.4 \\ \hline
\hline

\textbf{MSVAMP} & \textbf{Avg.} & \textbf{Lrl.} & \textbf{Hrl.} & \textbf{Bn} & \textbf{Th} & \textbf{Sw} & \textbf{Ja} & \textbf{Zh} & \textbf{De} & \textbf{Fr} & \textbf{Ru} & \textbf{Es} & \textbf{En} \\ \hline
MetaMath & 47.5 & {15.5} & \textbf{61.2} & 13.5 & 16.1 & 16.9 & 53.9 & 56.2 & \textbf{63.7} & \textbf{64.9} & 57.8 & \textbf{64.6} & \textbf{67.6} \\
LangBridge-En & 52.0 & {45.1} & {54.9} & 46.8 & 46.3 & 42.1 & 45.5 & 50.4 & 58.1 & 57.0 & 55.8 & 56.9 & 60.6 \\
Translate & 49.0 & {44.5} & {50.9} & 49.3 & 44.2 & 40.1 & 42.0 & 48.0 & 46.5 & 49.5 & 45.1 & 57.9 & 67.6 \\ \hline
MetaMath-Mul & 37.8 & {33.7} & {39.6} & 30.0 & 35.1 & 36.0 & 38.3 & 37.9 & 39.1 & 41.4 & 39.0 & 41.7 & 39.8 \\
MathOctopus & 38.1 & {33.1} & {40.3} & 27.3 & 32.9 & 39.1 & 39.2 & 38.2 & 40.1 & 43.2 & 38.8 & 41.4 & 41.1 \\
LangBridge & 54.4 & {51.6} & {55.5} & 49.9 & 52.2 & 52.7 & 53.3 & 54.1 & 56.0 & 56.4 & 54.7 & 56.1 & 58.1 \\
MindMerger & {58.0} & {53.1} & {60.2} & \textbf{52.2} & {53.2} & {53.9} & {57.3} & {57.0} & {61.3} & {60.2} & {58.1} & {62.9} & {64.3} \\
\textbf{\mname} & \textbf{59.1} & \textbf{54.6} & {61.1} & 51.8 & \textbf{55.1} & \textbf{56.9} & \textbf{59.3} & \textbf{58.7} & 62.5 & 62.1 & \textbf{58.8} & 62.0 & 64.0 \\ \hline
\end{tabular}
}
\caption{\label{tab:mgsm}
    Experimental results on MGSM and MSVAMP datasets. `Lrl.', `Hrl.', and `Avg.' represent the average accuracy across low-resource languages, high-resource languages, and all languages, respectively. Referring to \citet{mindmerger}, we regard Bn, Th, and Sw as low-resource languages, and regard the remaining languages as high-resource languages. Models above the line are trained in English, while those below are trained in multiple languages. The languages corresponding to the abbreviations used in the tables are provided in Appendix \ref{sec:appendix_com_eva}.
  }
\end{table*}

Table \ref{tab:mgsm} presents the results of the mathematical reasoning tasks. As shown, \mname significantly surpasses all baselines, outperforming the current state-of-the-art, MindMerger, by 1.6\% on MGSM and 1.1\% on MSVAMP in terms of average accuracy across all languages. These results highlight the effectiveness of \mname on English LLM.

For methods that directly finetune LLMs, such as MetaMath, MetaMath-Mul, and MathOctopus, it is challenging to achieve strong performance across both high-resource and low-resource languages simultaneously. Training exclusively in English (e.g., MetaMath) generally results in high performance for high-resource languages like English, but poor results in low-resource languages. Conversely, methods trained on multilingual data (e.g., MetaMath-Mul and MathOctopus) often suffer from a significant performance drop in high-resource languages. For instance, MetaMath-Mul's performance declines by 11.8 and 21.6 points on high-resource languages in the MGSM and MSVAMP datasets, respectively. This demonstrates the difficulty of achieving consistently high performance across both high-resource and low-resource languages in LLM-based models.

This challenge can be significantly alleviated by models such as LangBridge, MindMerger, and LayAlign, which share a common architecture that integrates a multilingual encoder with LLMs. All three models demonstrate substantial improvements over traditional LLM-based approaches in both high- and low-resource languages. Among them, \mname achieves the best performance on the MGSM and MSVAMP benchmarks, highlighting its ability to effectively leverage the representations from the multilingual encoder through the layer-wise aligner and \attaname mechanisms.

We further compare \mname with Translate. \mname surpasses Translate by a substantial margin, showing improvements of 15.9 points on MGSM and 10.1 points on MSVAMP. Additional, Translate suffers from longer inference times and reliance on external translation systems due to its need to translate multilingual prompts into English.

\subsection{Commonsense Reasoning and Language Understanding}
\subsubsection{Experimental Setup}
\paragraph{Evaluation Datasets.} We evaluate commonsense reasoning and language understanding capabilities using X-CSQA \cite{xcsqa} and XNLI \cite{xnli}, respectively.

\paragraph{Training Datasets.} For both tasks, we adopt the same dataset setup as MindMerger \cite{mindmerger}. The Lego-MT translation dataset \cite{lego-mt} is utilized in the first training stage, while the translated X-CSQA training set \cite{clues,mindmerger} and the official development set of XNLI are used in the second training stage for the commonsense reasoning and language understanding tasks, respectively.

\paragraph{Baselines.} We compare \mname with two LLM-based baselines:

\noindent $\bullet$ \textbf{LLaMAX2} \cite{llamax-xcsqa}: fine-tuned from the powerfull multilingual model LLaMAX2-7B on an English task dataset, with LLaMAX2-7B covering all the languages examined in this study.
We utilize the official checkpoints.\footnote{Commonsense reasoning: \url{https://huggingface.co/LLaMAX/LLaMAX2-7B-X-CSQA}; Language understanding: \url{https://huggingface.co/LLaMAX/LLaMAX2-7B-XNLI}}. 

\noindent $\bullet$ \textbf{LLaMAX2-Mul}: fine-tuned from LLaMAX2 using the same multilingual task dataset as ours. 

We also include LangBridge and MindMerger in our comparisons. To ensure a fair evaluation, all models, including LangBridge, MindMerger, and \mname, utilize LLaMAX2 as the LLM and \texttt{mT5-xl} encoder as the multilingual encoder, with all models trained on the same two-stage dataset.

\subsubsection{Results}
\begin{table}[t]
\centering
  \resizebox{0.85\columnwidth}{!}{
  \begin{tabular}{l|c|ccc}
    \hline
    X-CSQA & Avg. & Sw & Fr & En \\
    \hline
    LLaMAX2 & 55.0 & 43.1 & 61.4 & 73.9 \\
    LLaMAX2-Mul & 49.4 & 39.2 & 53.4 & 68.6 \\
    LangBridge & 56.7 & 52.5 & 60.4 & 62.4 \\ 
    MindMerger & 61.2 & 51.5 & 64.5 & 75.6 \\
    \mname & \textbf{62.3} & \textbf{53.3} & \textbf{66.5} & \textbf{76.7} \\ \hline
    \hline
    XNLI & Avg. & Sw & Fr & En \\
    \hline
    LLaMAX2 & 76.5 & 66.7 & 83.1 & \textbf{89.7} \\
    LLaMAX2-Mul & 77.4 & 68.3 & \textbf{84.7} & 89.3 \\
    LangBridge & 76.0 & 72.2 & 78.0 & 80.8 \\ 
    MindMerger & 79.2 & 72.7 & 84.2 & 88.5 \\
    \mname & \textbf{79.7} & \textbf{73.0} & \textbf{84.7} & 88.9 \\ \hline    
  \end{tabular}
  }
  \caption{Experimental results on X-CSQA and XNLI datasets. Due to limited space, we list several representative languages in this table. The complete results is in Table \ref{tab:c_xcsqa} and Table \ref{tab:c-xnli} of Appendix \ref{sec:appendix_com_eva}.}
  \vspace{-10pt}
  \label{tab:xcsqa_xnli}  
\end{table}

The results for X-CSQA and XNLI are presented in Table \ref{tab:xcsqa_xnli}. As shown, \mname sets a new state-of-the-art, outperforming LangBridge by 9.9\% and MindMerger by 1.8\% on X-CSQA, and improving by 4.9\% and 0.6\%, respectively, on XNLI. These results demonstrate that \mname is effective not only on English LLM backbones but also on multilingual LLM backbones.

Since the LLM backbone is inherently multilingual across all languages tested, fine-tuning it on English task datasets already yields strong multilingual task performance. For example, LLaMAX2 achieves scores of 55.0 on X-CSQA and 76.5 on XNLI. This makes further improvements challenging, as both LLaMAX2-Mul, which is fine-tuned on the multilingual task data, and LangBridge, which integrates a multilingual encoder into LLaMAX2, show only marginal gains or even performance declines. In contrast, \mname delivers robust performance in both commonsense reasoning and language understanding, underscoring the effectiveness of the layer-wise aligner, adaptive fusion-enhanced attention, and LLM text input integration.

\subsection{Ablation Studies}

\begin{table}
\centering
\resizebox{0.9\columnwidth}{!}{
\begin{tabular}{l|rrr}
\hline
MGSM & Avg. & Lrl. & Hrl. \\ \hline
w/o Adapter & 44.1 & 15.9 & 56.2 \\
w/o LLM Input & 56.8 & 55.5 & 57.3 \\
w/o Layer-Wise Aligner & 56.9 & 53.1 & 58.6 \\ \hline
w/o Translation Stage & 52.0 & 38.9 & 57.5 \\
w/o Task Stage & 38.8 & 24.7 & 44.9 \\ \hline
MetaMath & 37.9 & 5.9 & 51.6 \\
LayAlign & 59.0 & 56.4 & 60.2 \\ \hline
\end{tabular}
}
\caption{\label{tab:ablation}
    Ablation experiments of \mname on the MGSM dataset. The complete table of accuracy for each language is in Table \ref{tab:c-ablation}.
  }
\vspace{-10pt}
\end{table}

We conduct ablation studies to examine the contributions of key components in our method, including the adapter, the layer-wise aligner, the LLM input embedding $I_\mathrm{in}$, and the two-stage training approach. Table \ref{tab:ablation} presents the ablation results. 
Note that the layer-wise aligner and \attaname operate in conjunction, so removing the layer-wise aligner also disables \attaname. As shown in Table \ref{tab:ablation}, all components significantly contribute to \mname's overall performance. Since the adapter receives the highest-level representations from the encoder, its removal results in a substantial performance drop of 7.0 points. The task fine-tuning stage, which is directly related to downstream evaluations, also plays a critical role in the model's success.
The layer-wise aligner and the translation stage are all integral to \mname, with their absence leading to performance declines of 2.1, 2.2, and 7.0 points, respectively. Notably, even without task-specific fine-tuning, \mname outperforms MetaMath by 33.0 points on low-resource languages, demonstrating that aligning the multilingual encoder with the LLM enhances task performance in low-resource settings, even in the absence of specialized training.
To evaluate the role of the gating mechanism in \mname, we also conduct an ablation study by removing the gate. Without the gate, we observe that the training loss of \mname fails to decrease effectively. This highlights the gating mechanism's critical role in ensuring smooth and stable training.

\section{Analyses}\label{sec:analyses}

\subsection{Multilingual Encoder}
\begin{table}
\centering
\resizebox{0.8\columnwidth}{!}{
\begin{tabular}{l|l|lll}
\hline
MGSM & Parm(M) & Avg. & Lrl. & Hrl. \\ \hline
mGPT & 1418 & 48.5 & 30.8 & 56.1 \\
XGLM & 1733 & 51.1 & 42.4 & 54.8 \\ \hline
NLLB & 1733 & 55.3 & 50.8 & 57.2 \\
mT5-xl & 1670 & 59.0 & 56.4 & 60.2 \\ \hline
\end{tabular}
}
\caption{\label{tab:encoder_table}
    Experiments on MGSM using MetaMath as LLM and different multilingual models as encoders. The complete table for each language is in Table \ref{tab:c-encoder}.
  }
\vspace{-10pt}
\end{table}

The \mname framework allows for the flexible selection of various multilingual models as encoders to extract multilingual representations. We evaluated several multilingual models on the MGSM benchmark, including the encoder from the two encoder-decoder multilingual models \texttt{mT5-xl} \cite{mT5} and \texttt{NLLB-200-3.3B} \cite{nllb}, as well as the decoder-only architectures mGPT \cite{mgpt} and XGLM \cite{xglm}. As shown in Table \ref{tab:encoder_table}, the encoder from \texttt{mT5-xl} achieves the best performance, while the encoders from the encoder-decoder multilingual models generally outperform those using multilingual decoders as \mname's encoder.

\subsection{Training on English Task Data} \label{sec:eff_llm_input}

In prior experiments, we demonstrated the effectiveness of \mname under multilingual training conditions. However, obtaining task-specific data for low-resource languages remains a significant challenge. To address this, we examine the performance of \mname when trained exclusively on English task-specific data by replacing the task-stage training set with the English MetaMath-200k dataset.
Since both the input and output are in English, the LLM input could act as a shortcut for the model, potentially harming the learning of the multilingual aligner and adapter during finetuning. This may lead to poor performance in low-resource languages. Conversely, removing the LLM input text forces the model to depend on the multilingual encoder, encouraging cross-lingual generalization.
To verify this, we evaluate three variants of \mname on the MGSM benchmark: the full \mname model, \mname without the LLM input text $I_\mathrm{in}$, and LangBridge, which serves as a baseline equivalent to \mname without both the LLM input and the layer-wise aligner.

As shown in Table~\ref{tab:english_training}, when trained in an English-only setting, \mname tends to exploit the shortcut by relying heavily on the English LLM input. As a result, the multilingual information from the encoder is largely ignored during finetuning, leading to poor performance on low-resource languages.
In contrast, the \mname variant without LLM input text is forced to rely on the multilingual information provided by the \texttt{mT5} encoder during finetuning. The superior performance of this variant underscores the critical importance of the layer-wise aligner, particularly in English-only downstream finetuning. In this setting, the \mname variant without LLM input text is recommended to enhance the model's multilingual capabilities, as it effectively leverages the multilingual encoder for improved cross-lingual generalization.

\begin{table}[]
\centering
\resizebox{0.85\columnwidth}{!}{
\begin{tabular}{lccc}
\hline
\multicolumn{1}{l}{MGSM} & Avg. & Lrl. & Hrl. \\ \hline
\multicolumn{1}{l}{LangBridge} & 49.1 & 44.4 & 51.1 \\
\multicolumn{1}{l}{\mname w/o LLM Input} & 51.8 & 45.7 & 54.5 \\ 
\multicolumn{1}{l}{\mname} & 38.1 & 5.3 & 52.1 \\ \hline
\end{tabular}
}
\caption{\label{tab:english_training}
    Experiments on MGSM using English-only task data. The complete table of accuracy for each language is in Table \ref{tab:c-english}.
  }
\vspace{-10pt}
\end{table}

\subsection{Empowering Multilingual LLM for Low-Resource Languages}

\begin{figure}[t]
  \includegraphics[width=\linewidth]{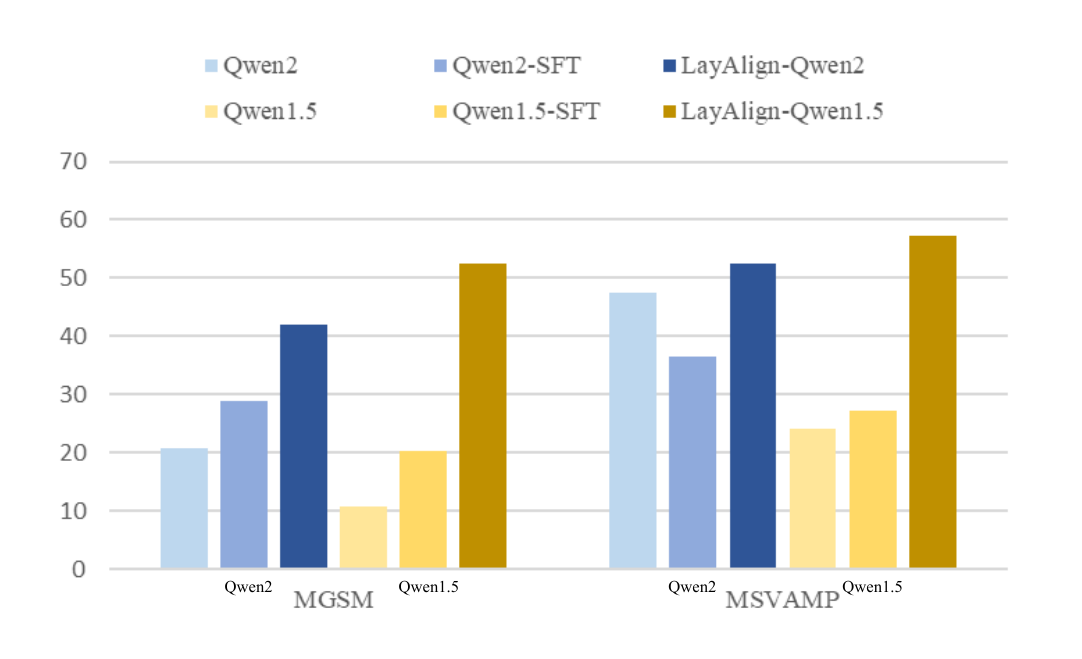}
  \caption{Experimental results for the Swahili language on the MGSM and MSVAMP datasets.}
  \label{fig:qwen_math}
\vspace{-10pt}
\end{figure}

In Section~\ref{sec:experiments}, we applied \mname to both the English-centric LLaMA2 backbone and its multilingual variant, LLaMAX2, which supports all languages evaluated. Here, we further investigate whether \mname can empower multilingual LLMs to improve performance in low-resource languages where they underperform. To this end, we utilize the advanced LLMs Qwen1.5-7B-Chat \cite{qwen1.5} and Qwen2-7B-Instruct  \cite{qwen2}, which exhibit strong multilingual capabilities but face challenges in scaling to less-represented languages in their training data. We conduct experiments on the MGSM and MSVAMP benchmarks, focusing on Swahili (Sw), a less-represented language.

Figure~\ref{fig:qwen_math} presents the results, comparing \mname with the vanilla Qwen models and Qwen-SFT fine-tuned on the same multilingual mathematical dataset used in our study. As shown, \mname consistently outperforms the baseline methods on both the MGSM and MSVAMP tasks. Comparing vanilla Qwen and Qwen-SFT, we observe that directly fine-tuning these LLMs on multilingual mathematical datasets containing Swahili yields only marginal improvements and, in some cases, even degrades performance. In contrast, \mname significantly boosts model performance. On MGSM, \mname improves Qwen1.5 and Qwen2 by 41.6 and 21.2 points, respectively, while on MSVAMP, it enhances their performance by 33.2 and 5.1 points, respectively. These results further underscore the potential of our method.

\subsection{Analyses of Representation Space}
\label{sec:analysis_representation}
\begin{figure}[]
  \includegraphics[width=\linewidth]{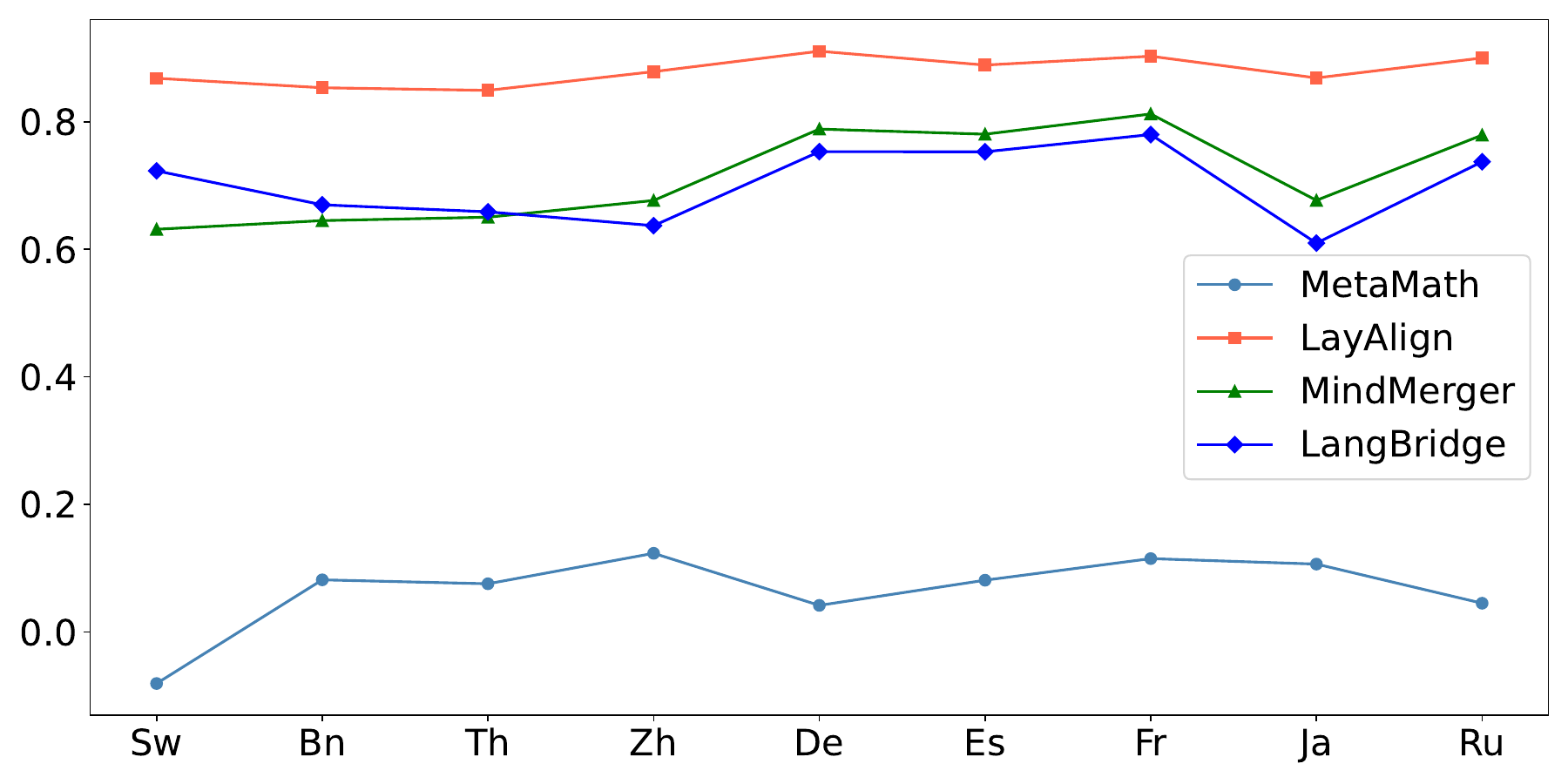}
  \caption{The cosine similarities of the final layer of LLM pooled output representations of English with other languages obtained with the FLORES-101 dataset.}
  \label{fig:cos_sim}
\vspace{-15pt}
\end{figure}

\begin{figure}[t]
  \includegraphics[width=\linewidth]{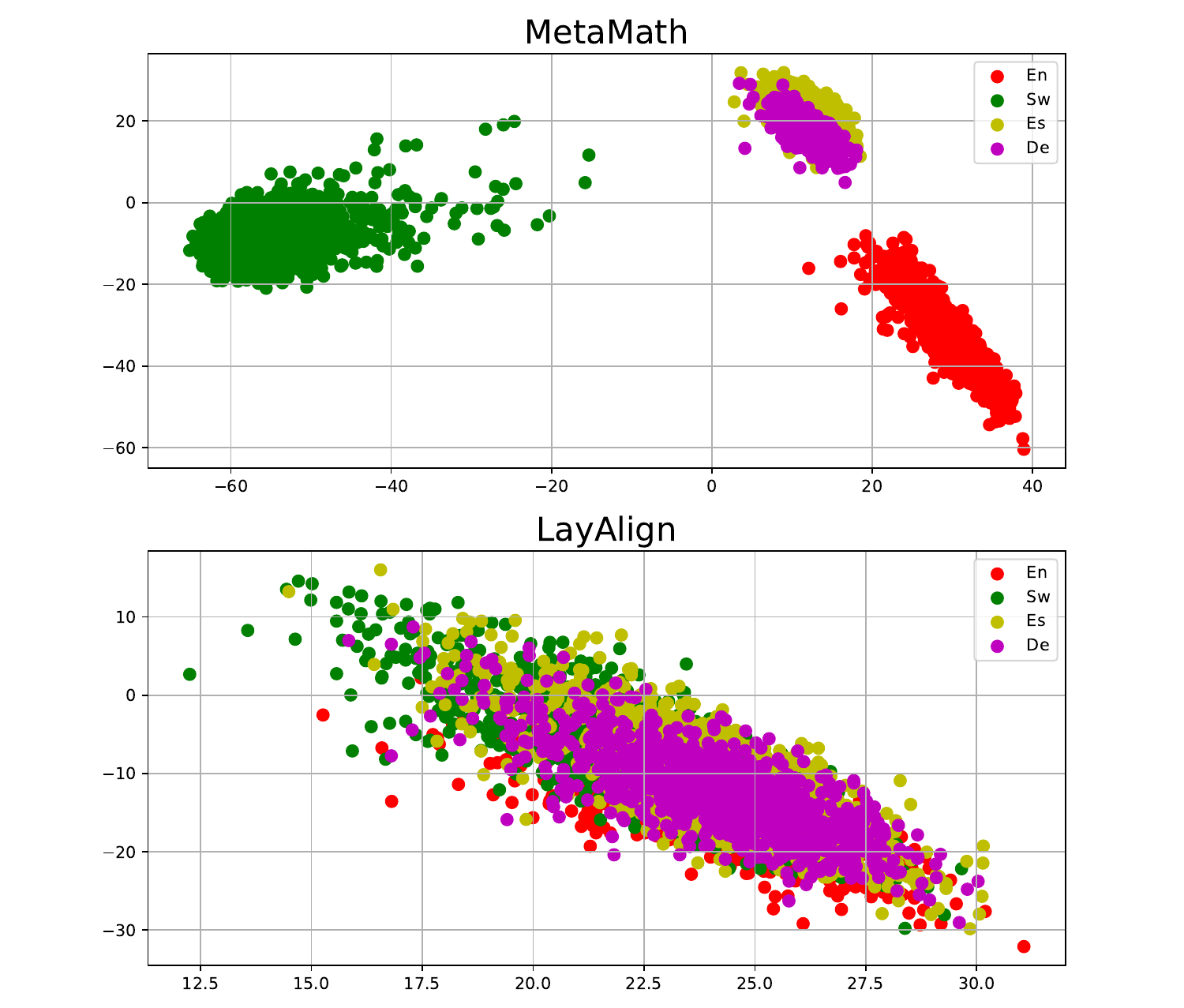}
  \caption{First two principal components of pooled output representations obtained with the FLORES-101.}
  \label{fig:pca}
\vspace{-10pt}
\end{figure}

To evaluate whether \mname effectively aligns multilingual representations, we compute the cosine similarity between the mean-pooled representations of English and other languages in MGSM, such as Chinese (Zh) and Swahili (Sw), from the final layer of the LLM using the FLORES-101 dataset \cite{flores}. Figure \ref{fig:cos_sim} presents the results for different methods, clearly demonstrating that \mname achieves more effective alignment of representations across languages compared to baseline methods. This alignment contributes to the superior performance of \mname.

We further illustrate this by visualizing the representations of \mname and MetaMath using Principal Component Analysis, as shown in Figure \ref{fig:pca}. For MetaMath, high-resource languages like Spanish (Es) and German (De) align closely with English (En), while low-resource languages like Swahili (Sw) are positioned much farther from English. In contrast, \mname unifies all languages into a single cluster, indicating more effective alignment of multilingual representations.

\section{Conclusion}
In this paper, we introduce \mname, a simple yet effective method designed to leverage multilingual encoders for enhancing the multilingual reasoning capabilities of LLMs. We demonstrate that our approach yields consistent improvements over existing baselines. Notably, \mname shows effectiveness in improving cross-lingual reasoning when trained on English-only task data, and \mname enables multilingual LLMs to scale to less-represented languages in their training data. Additionally, we provide analyses indicating that \mname aligns the representations of various languages with English more effectively. We hope these findings will benefit low-resource language users and inspire further research in this field.


\section*{Limitations}
While \mname can enhance the performance of English-centric LLMs in low-resource languages through multilingual task training, there remains a performance gap compared to models specifically pretrained and fine-tuned in the target languages. 

\bibliography{custom}

\appendix

\section{Additional Analysis Experiments}
\subsection{Analysis of Representation across Layers}

\begin{figure}[htbp]
	\centering
	\begin{subfigure}{0.99\linewidth}
		\centering
		\includegraphics[width=0.9\linewidth]{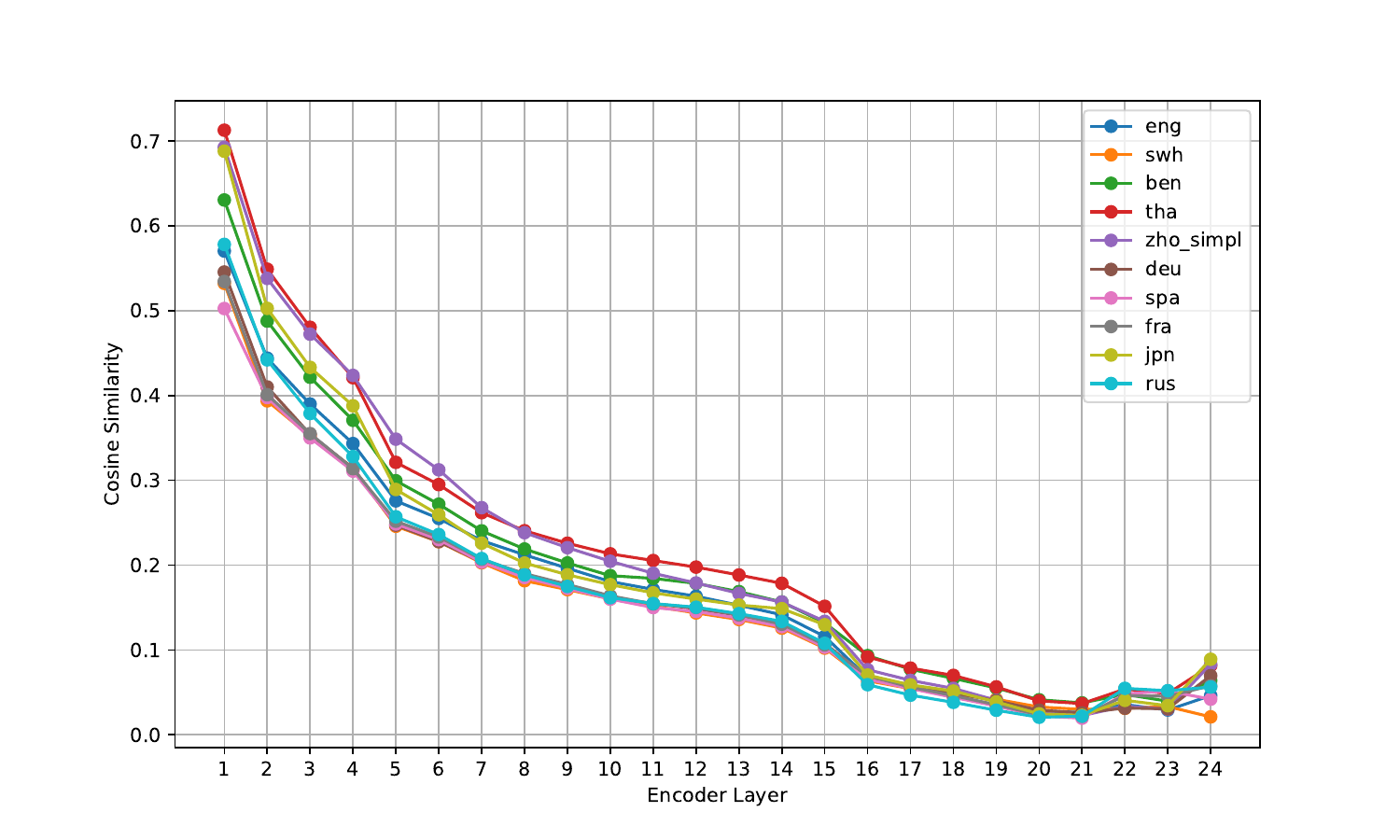}
		\caption{Embedding Layer}
		\label{Embedding Layer}
	\end{subfigure}
	\centering
	\begin{subfigure}{0.99\linewidth}
		\centering
		\includegraphics[width=0.9\linewidth]{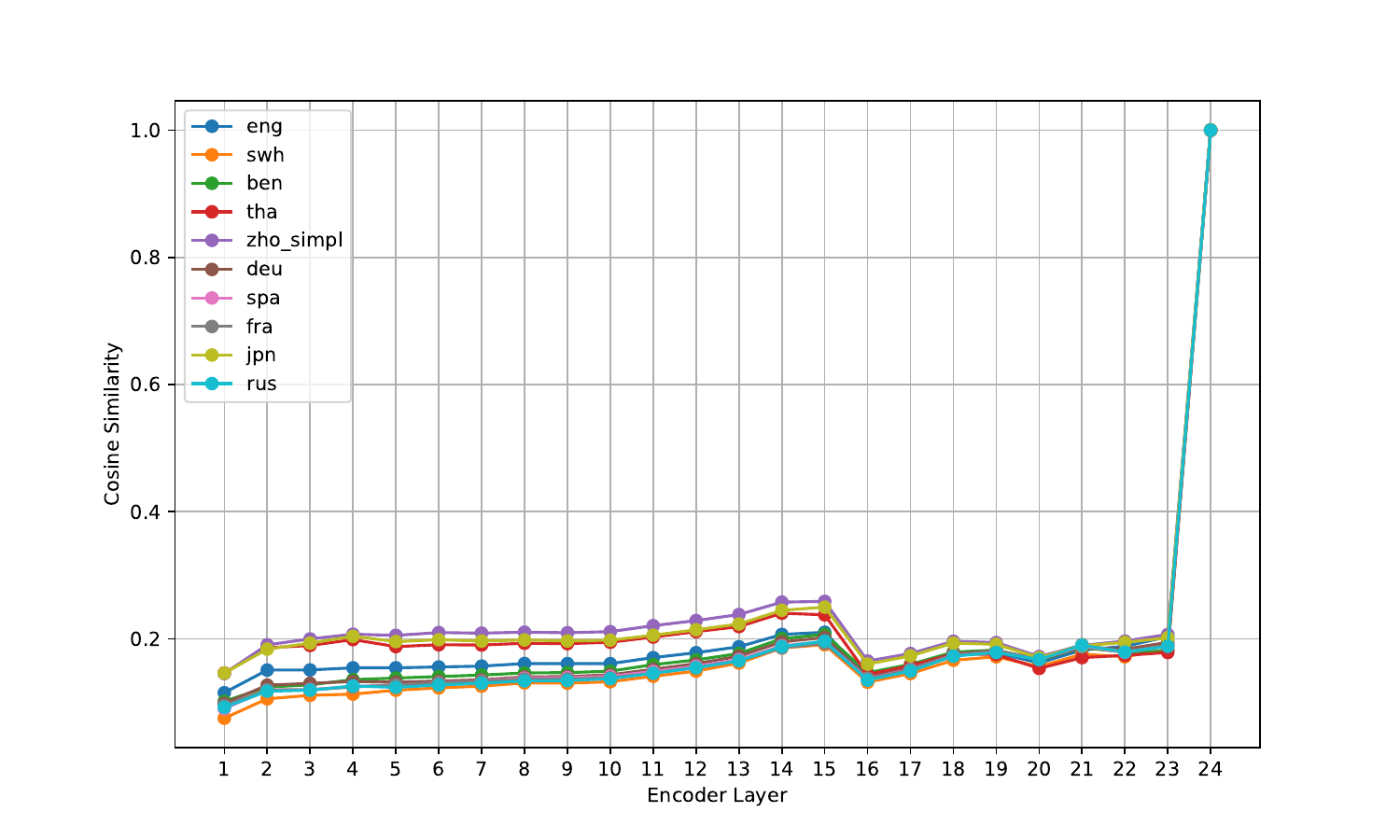}
		\caption{Last Layer}
		\label{Last Layer}
	\end{subfigure}
        \caption{Cosine similarity. (a) Cosine similarity between the $i$-th encoder layer and the embedding layer in mT5, utilizing the FLORES-101 dataset. (b) Cosine similarity between the $i$-th encoder layer and the last layer in mT5, utilizing the FLORES-101 dataset.}
  \label{fig:cosine_sim}
\end{figure}

\begin{figure*}[htbp]
\centering
\begin{subfigure}{0.325\linewidth}
    \centering
    \includegraphics[width=0.9\linewidth]{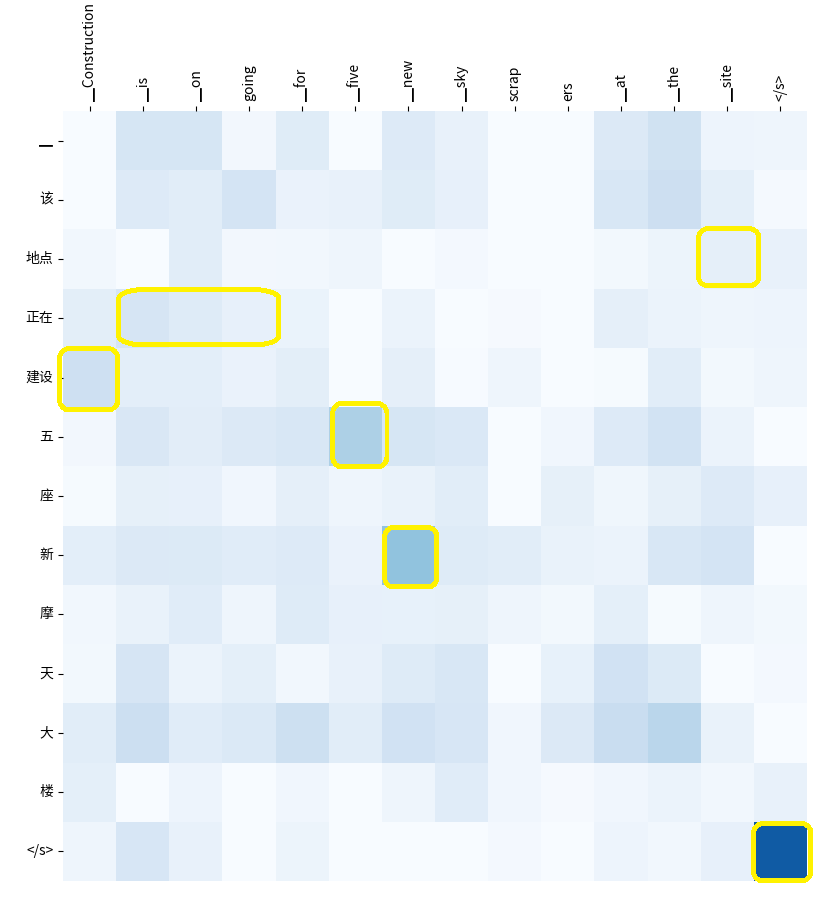}
    \caption{1th layer of Encoder}
    \label{1th}
\end{subfigure}
\centering
\begin{subfigure}{0.325\linewidth}
    \centering
    \includegraphics[width=0.9\linewidth]{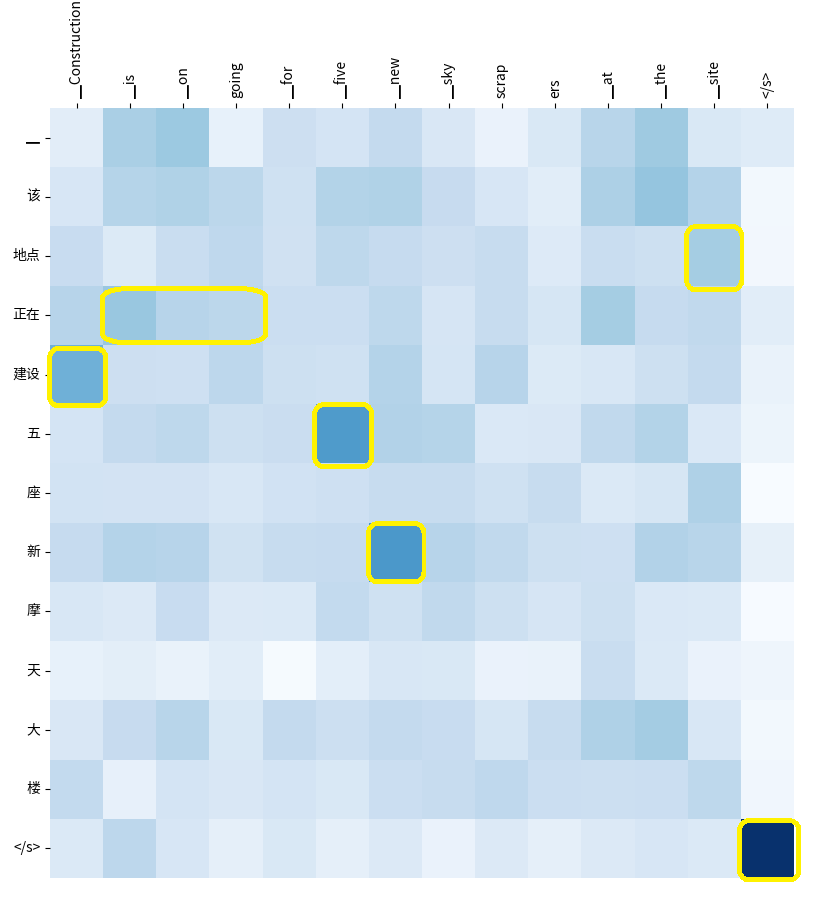}
    \caption{12th layer of Encoder}
    \label{12th}
\end{subfigure}
\centering
\begin{subfigure}{0.325\linewidth}
    \centering
    \includegraphics[width=0.9\linewidth]{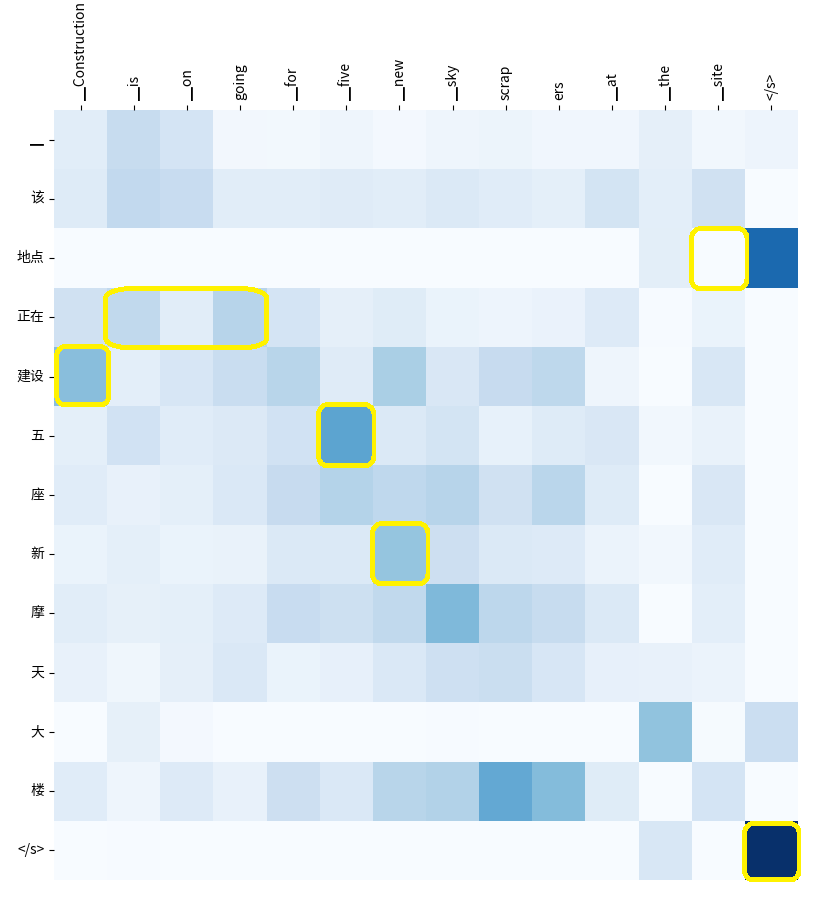}
    \caption{24th layer of Encoder}
    \label{24th}
\end{subfigure}
\caption{Visualization of the cosine similarity between representations of Chinese and English sentences in the Encoder.}
\label{fig:encoder_representation}
\end{figure*}
Each encoder layer's representation has different levels of granularity information. As the depth of the encoder layers increases, each layer produces increasingly coarse-grained descriptions of the global context \cite{Layer-Wise-2}. As shown in Figure \ref{fig:cosine_sim}, the cosine similarity between the final layer and other encoder layers is markedly different, and the cosine similarity between the $i$-th encoder layer and the embedding layer decreases as $i$ increases. This reflects the shifting granularity of information across different encoder layers. Furthermore, prior studies suggest that intermediate layers can be seen as noisy versions of the final layer’s representation, improving model robustness when using layer-wise representations \cite{Layer-Wise-1}. Therefore, we use the layer-wise representation of the multilingual encoder to better utilize the language understanding of the encoder to improve the multilingual capabilities of LLM.

To further analyze the representation of different encoder layers, Figure \ref{fig:encoder_representation} shows the cosine similarity of representations for Chinese and English tokens across the first, twelfth, and final encoder layers. In the first encoder layer, cosine similarity is relatively low, with only token pairs like `五' and `five,' and `新' and `new' showing better alignment. By the twenty-fourth layer, many tokens become aligned, yet the similarity between `新' and `new,' and `地点' and `site,' is lower than in the twelfth layer. This suggests that the twelfth layer can provide alignment information that supports the final layer. Therefore, utilizing layer-wise representations is crucial for fully leveraging the multilingual capabilities of the encoder.

\begin{table*}[t]
    \resizebox{\textwidth}{!}{
\begin{tabular}{l|ccc|cccccccccc}
\hline
\textbf{MGSM} & \textbf{Avg.} & \textbf{Lrl.} & \textbf{Hrl.} & \textbf{Bn} & \textbf{Th} & \textbf{Sw} & \textbf{Ja} & \textbf{Zh} & \textbf{De} & \textbf{Fr} & \textbf{Ru} & \textbf{Es} & \textbf{En} \\ \hline
\mname-LR & 57.5 & {52.9} & {59.5} & 48.8 & 54.0 & 56.0 & 53.2 & 55.6 & 59.6 & 62.0 & 58.8 & 60.4 & {66.8} \\
{\mname} & {59.0} & {56.4} & {60.2} & {51.6} & {59.2} & {58.4} & {52.0} & {56.0} & {62.0} & {61.6} & 61.6 & {61.6} & 66.4 \\ \hline

\end{tabular}
}
\caption{\label{tab:LR}
    Experimental results of \mname and \mname-LR on the MGSM dataset. \mname-LR refers to the variant of \mname where only the final layer representation from the multilingual encoder is fed into the LLM's \attaname module.
  }
\end{table*}

\begin{table*}[t]
    \resizebox{\textwidth}{!}{
\begin{tabular}{l|ccc|cccccccccc}
\hline
\textbf{MGSM} & \textbf{Avg.} & \textbf{Lrl.} & \textbf{Hrl.} & \textbf{Bn} & \textbf{Th} & \textbf{Sw} & \textbf{Ja} & \textbf{Zh} & \textbf{De} & \textbf{Fr} & \textbf{Ru} & \textbf{Es} & \textbf{En} \\ \hline
\mname (Dgate) & 58.8 & {54.5} & {60.6} & 49.6 & 55.6 & 58.4 & 54.4 & 56.4 & 59.2 & 62.0 & 61.2 & 64.4 & {66.8} \\
{\mname} & {59.0} & {56.4} & {60.2} & {51.6} & {59.2} & {58.4} & {52.0} & {56.0} & {62.0} & {61.6} & 61.6 & {61.6} & 66.4 \\ \bottomrule
\end{tabular}
}
\caption{\label{tab:gate_mgsm}
    Experimental results of LayAlign with different gating mechanisms on the MGSM dataset. Dgate denotes dynamic gate.
  }
\end{table*}

\begin{table}[h]
    \resizebox{\columnwidth}{!}{
    \begin{tabular}{lccccc}
        \toprule
        \textbf{Settings} & \textbf{User Input in Stage 1} & \textbf{User Input in Stage 2} & \textbf{Avg.} & \textbf{Lrl.} & \textbf{Hrl.} \\
        \midrule
        \multicolumn{6}{c}{Multilingual Task Data} \\

        LayAlign + LLM Input at Trans & $\surd$  & $\surd$  & 57.5 & 50.5 & 60.5 \\
        LayAlign - LLM Input at Task & $\times$  & $\times$  & 56.8 & 55.5 & 57.3 \\
        LayAlign & $\times$  & $\surd$  & \textbf{59.0} & \textbf{56.4} & \textbf{60.2} \\
        \midrule
        \multicolumn{6}{c}{English Task Data} \\

        LayAlign + LLM Input at Trans & $\surd$ & $\surd$ & 37.8 & 6.3 & 51.4 \\
        LayAlign - LLM Input at Task & $\times$ & $\times$ & \textbf{51.8} & \textbf{45.7} & \textbf{54.5} \\
        LayAlign & $\times$ & $\surd$ & 38.1 & 5.3 & 52.1 \\
        \bottomrule
    \end{tabular}
    }
    \caption{Experiments of user input at different stages on MGSM. `Lrl.', `Hrl.', and `Avg.' represent the average accuracy across low-resource languages, high-resource languages, and all languages, respectively.}
    \label{tab:w1_input_ablation}
\end{table}

To further assess whether utilizing representations from all layers of the multilingual encoder can enhance the model's multilingual capabilities, we fed the final layer's representation from the multilingual encoder into LayAlign’s \attaname. We conducted experiments on the MGSM dataset, and the results are presented in Table \ref{tab:LR}. Compared to using only the final layer representation from the multilingual encoder, LayAlign achieved a 1.5\% improvement, indicating that representations from other layers of the multilingual encoder also contribute to enhancing the model's multilingual performance.

\subsection{Analysis of Adaptive Fusion-Enhanced Attention}
\begin{figure}[h]
  \includegraphics[width=\linewidth]{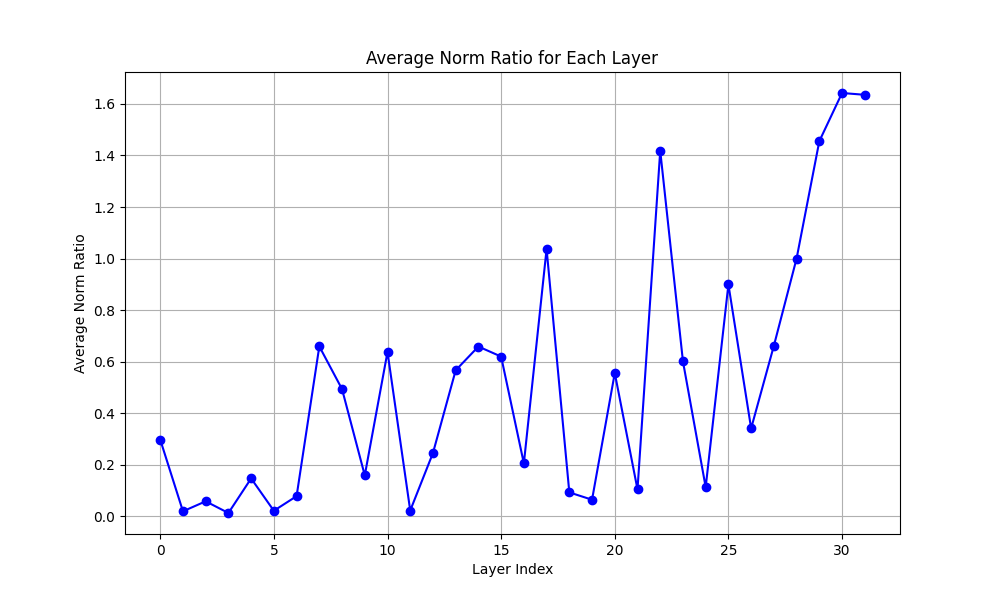}
  \caption{Norm ratio between gate-weighted cross-attention and self-attention across LLM layers using the FLORES-101 dataset. Cross-attention shows a stronger effect in deeper layers.}
  \label{fig:norm_ratio}
\end{figure}

\begin{figure}[t]
  \includegraphics[width=\linewidth]{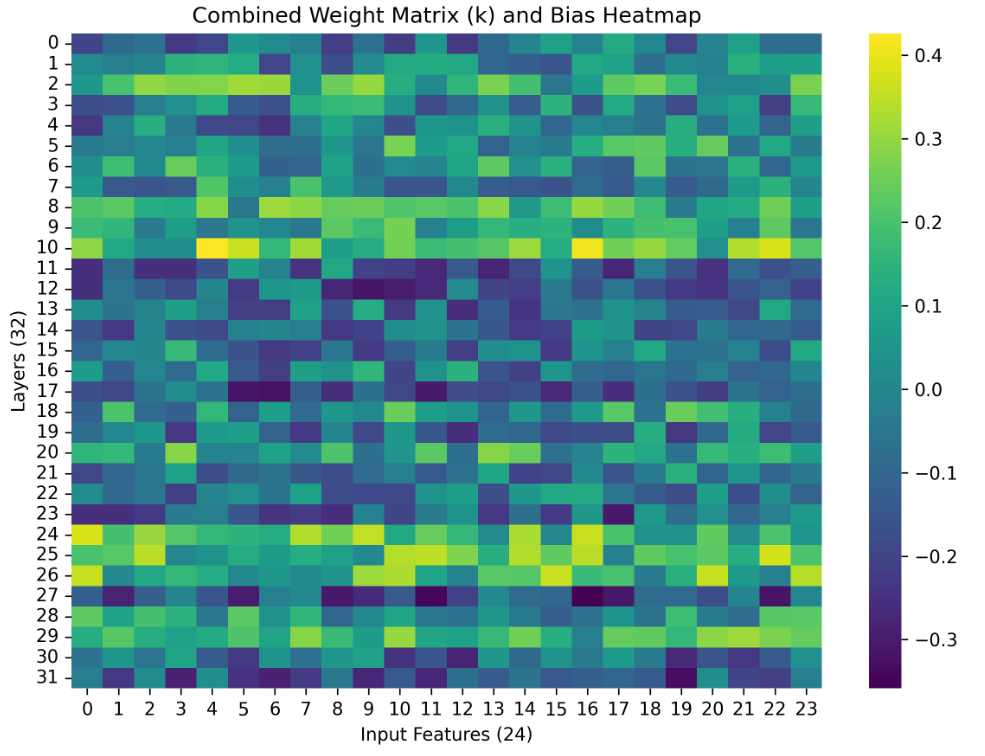}
  \caption{Visualization of the layer-wise aligner}
  \label{fig:vis_aligner}
\end{figure}

To validate the impact of cross-attention in the \attaname mechanism, we compute the ratio between the norm of the gate-weighted cross-attention and that of the self-attention across all layers of the LLM, utilizing the FLORES-101 dataset for visualization. As shown in Figure \ref{fig:norm_ratio}, the gate-weighted cross-attention significantly influences the overall attention mechanism, with a more pronounced effect in the deeper layers of the LLM. 

We also visualize the layer-wise aligner, as shown in Figure \ref{fig:vis_aligner}. The layer-wise aligner effectively integrates the representations from different layers of the multilingual encoder, providing the LLM with enriched multilingual information by leveraging these fused representations.

\mname implements a layer-wise fusion gate that is independent of the current hidden state. To validate this approach, we compare it with a variant called the dynamic gate, where the gate at each layer is determined by the current hidden state. As shown in Table \ref{tab:gate_mgsm}, while the dynamic gate yields competitive results, it slightly underperforms compared to the fusion gate in \mname. Moreover, our fusion gate requires fewer gate parameters to be trained, as each layer only requires a single scalar gate parameter.

\subsection{The Analysis of User's Input Text}
In our approach, during the translation stage, we use only the output from the adapter as input to the LLM, excluding any user text input. To assess the impact of user text as LLM input during this stage, we conduct experiments on the MGSM dataset. The results are presented in Table \ref{tab:w1_input_ablation}. 
\paragraph{Stage 1: Translation Stage}  
The primary objective of Stage 1 is to align the representation space of the multilingual encoder to the LLM through translation-based alignment. As shown in Table \ref{tab:w1_input_ablation}, it is consistently more effective to omit LLM text input during the translation stage rather than include it.

\paragraph{Stage 2: Task Stage}  
In contrast, Stage 2 focuses on leveraging both the multilingual encoder and the LLM for task-specific reasoning. \textbf{In multilingual tasks,} including the user’s input text in addition to the adapter’s output maximizes the LLM’s reasoning potential. This configuration is particularly effective for high-resource languages, as the LLM benefits from its existing high-resource knowledge.
However, \textbf{for English tasks,} including user input in Stage 2 can act as a shortcut, causing the model to rely excessively on the LLM’s inherent English capabilities while neglecting the multilingual encoder. This is reflected in the performance drop observed when user input is included in Stage 2 (e.g., 38.1\% compared to 51.8\%). The best results are achieved when the LLM is forced to rely on the multilingual encoder rather than directly leveraging its internal English representations.

These results further validate our design choices, demonstrating that LayAlign’s two-stage input strategy effectively balances alignment and reasoning, leading to superior multilingual performance.

\subsection{Analysis of Parameters}
\begin{table*}[t]
    \resizebox{\textwidth}{!}{
\begin{tabular}{lccccc}
\hline
\textbf{Model} & \textbf{Adapter's Parameters(M)} & \textbf{Layer-Wise Aligner's Parameters(M)} & \textbf{Total Train Parameters (M)} & \textbf{Avg.}  \\ \hline
\mname & 25.18 & {8.39} & {33.57} & 59.0 \\
{\mname+} & {25.18} & {12.59} & {37.77} & {58.4} \\
MindMerger & 25.18 & {0} & {25.18} & 57.4 \\
{MindMerger+} & {37.76} & {0} & {37.76} & {57.3} \\ \bottomrule
\end{tabular}
}
\caption{\label{tab:app_parameter}
    The performance and parameters of models comparison on MGSM. + denotes the experiments with increased parameters.
  }
\end{table*}

The adapter in LayAlign has 25.18M parameters, and the layer-wise aligner contributes 8.39M parameters, resulting in a total of 33.57M trainable parameters. This lightweight design ensures efficiency while maintaining competitive performance.
As shown in Figure \ref{tab:app_parameter}, our experiments demonstrate that lightweight aligners are sufficient for collecting and leveraging information from all encoder layers. Notably, our findings align with the results reported for LangBridge \cite{langbridge}, which observed better performance with a simpler Linear adapter compared to a more parameter-intensive MLP design on the XCOPA benchmark (76.6\% vs. 72.7\%, respectively). This indicates that merely increasing the number of parameters in the aligner does not always yield performance improvements.
To evaluate whether larger aligners could enhance performance, we experimented with a modified aligner design, increasing the parameter count from 8.39M to 12.59M by introducing an additional Linear(2048, 2048) layer and a SiLU activation function. However, this modification led to a slight performance drop, with the average accuracy on MGSM decreasing from 59.0 to 58.4. Similarly, increasing the parameters of MindMerger (e.g., from 25.18M to 37.76M) did not result in performance gains, as the accuracy dropped from 57.4 to 57.3. These findings suggest that merely increasing the number of parameters is not a guaranteed path to better performance.

\subsection{Contribution of different Encoder Layers}

\begin{table*}[t]
    \resizebox{\textwidth}{!}{
\begin{tabular}{l|ccc|cccccccccc}
\hline
\textbf{MGSM} & \textbf{Avg.} & \textbf{Lrl.} & \textbf{Hrl.} & \textbf{Bn} & \textbf{Th} & \textbf{Sw} & \textbf{Ja} & \textbf{Zh} & \textbf{De} & \textbf{Fr} & \textbf{Ru} & \textbf{Es} & \textbf{En} \\ \hline
Last Hidden States & 57.5 & {52.9} & {59.5} & 48.8 & 54.0 & 56.0 & 53.2 & 55.6 & 59.6 & 62.0 & 58.8 & 60.4 & {66.8} \\
Avgerage Hidden States & {56.7} & {52.5} & {58.5} & {49.6} & {50.4} & {57.6} & {49.2} & {52.4} & {60.4} & {58.0} & 58.4 & {64.4} & 66.8 \\
First 8 Layers & 58.4 & {54.8} & {60} & 51.6 & 54.8 & 58.0 & 51.6 & 55.6 & 60.8 & 59.2 & 60.8 & 63.6 & {68.4} \\
Middle 8 layers & {57.3} & {53.9} & {58.7} & 48.4 & {55.6} & 57.6 & {50.8} & {52.4} & {58.8} & {59.6} & {60.4} & {61.2} & {68.0} \\
Last 8 layers & 57.9 & {54.1} & {59.5} & 52.0 & 52.4 & 58.0 & 50.8 & 54.4 & 63.6 & 58.0 & 56.4 & 65.6 & {68.0} \\
{\mname} & {59.0} & {56.4} & {60.2} & {51.6} & {59.2} & {58.4} & {52.0} & {56.0} & {62.0} & {61.6} & 61.6 & {61.6} & 66.4 
\\ \bottomrule
\end{tabular}
}
\caption{\label{tab:app_encoder_layers}
     Performance comparison of different mT5 encoder layer selections as inputs to the aligner on MGSM. 
  }

\end{table*}

In this section, we analyze which layers of mT5 contribute most to the performance improvements observed with LayAlign. To investigate this, we conducted experiments with various configurations for the aligner’s input, exploring different ways of extracting representations from the mT5 encoder. Specifically, we tested using (1) the final hidden layer, (2) the mean of all hidden layers, (3) the first 8 layers, (4) the middle 8 layers, and (5) the last 8 layers. The results are presented in Table \ref{tab:app_encoder_layers}.

Using only the last hidden states of the mT5 encoder led to an average performance drop of 1.5 points compared to LayAlign. This suggests that leveraging hidden states from multiple layers, rather than relying solely on the final layer, enhances the model’s capacity to comprehend multilingual text. The final layer alone appears insufficient for capturing the diverse and hierarchical information encoded across all layers.

Similarly, employing the average hidden states across all layers resulted in a 2.3-point decline in performance compared to LayAlign. This indicates that treating all hidden states as equally important is suboptimal, as it fails to fully exploit the rich linguistic information embedded in the multilingual encoder. In contrast, LayAlign’s adaptive strategy, which dynamically learns individual layer-wise weights, enables the model to prioritize layers based on their relevance to the task. This adaptive weighting mechanism highlights the varying contributions of different layers in supporting multilingual reasoning.

Moreover, when using hidden states from the first 8 layers, middle 8 layers, and last 8 layers, we observed that the first 8 layers yielded the best performance, while the middle 8 layers performed the worst. This suggests that the middle layers of mT5 contribute relatively less information to the LLM. Since the LLM already integrates the final-layer representation of mT5 through the adapter, incorporating the first 8 layers in the aligner provides additional shallow-layer information, further enriching the LLM’s multilingual understanding.

\section{Complete Evaluation Results}
\label{sec:appendix_com_eva}

In this paper, we utilize the following languages, with their respective abbreviations in parentheses. For clarity and ease of reference, these abbreviations are used throughout the text: Bengali (Bn), Thai (Th), Swahili (Sw), Japanese (Ja), Chinese (Zh), German (De), French (Fr), Russian (Ru), Spanish (Es), English (En), Urdu (Ur), Hindi (Hi), Arabic (Ar), Vietnamese (Vi), Polish (Pl), Flemish (Nl), Italian (It), Portuguese (Pt), Turkish (Tr), Greek (El), and Bulgarian (Bg). 

Due to space limitations in the main text, the complete results for different languages are provided in this section. Table \ref{tab:c_xcsqa} presents the complete experimental results on the X-CSQA dataset, while Table \ref{tab:c-xnli} reports the results on the XNLI dataset. Table \ref{tab:c-encoder} illustrates the performance of LayAlign when using different multilingual models as encoders and MetaMath as the LLM on the MGSM dataset. Table \ref{tab:c-ablation} provides the ablation study results for LayAlign on the MGSM dataset. Finally, Table \ref{tab:c-english} shows the experimental results on MGSM using English-only task data.

\begin{table*}
\resizebox{\textwidth}{!}{
\begin{tabular}{l|l|cccccccccccccccc}
\hline
X-CQSA & Avg & Ur & Sw & Hi & Ar & Vi & Ja & Pl & Zh & Nl & Ru & It & De & Pt & Fr & Es & En \\ \hline
LLaMAX2-X-CSQA & 55.0 & 38.9 & 43.1 & 44.3 & 45.5 & 54.1 & 49.4 & 54.6 & 58.1 & 58.5 & 56.9 & 59.1 & 58.9 & 61.1 & 61.4 & 62.7 & 73.9 \\
LLaMAX2-X-CSQA-SFT & 49.4 & 35.4 & 39.2 & 40.0 & 37.8 & 44.0 & 43.9 & 51.8 & 50.5 & 52.9 & 48.7 & 55.8 & 56.1 & 55.1 & 53.4 & 56.6 & 68.6 \\
LangBridge-SFT & 56.7 & 50.6 & 52.5 & 51.6 & 53.6 & 56.4 & 53.1 & 57.6 & 57.8 & 58.2 & 56.0 & 59.6 & 59.2 & 58.8 & 60.4 & 59.4 & 62.4 \\
MindMerger & 61.2 & 50.5 & 51.5 & 51.1 & 54.1 & 60.7 & 55.8 & 64.1 & 64.4 & 64.6 & 61.0 & 64.5 & 64.2 & \textbf{65.5} & 64.5 & \textbf{67.8} & 75.6 \\
\mname & \textbf{62.3} & \textbf{51.7} & \textbf{53.3} & \textbf{53.7} & \textbf{55.9} & \textbf{62.0} & \textbf{56.4} & \textbf{64.8} & \textbf{64.6} & \textbf{66.2} & \textbf{62.0} & \textbf{66.2} & \textbf{65.2} & 64.3 & \textbf{66.5} & 67.3 & \textbf{76.7} \\ \hline
\end{tabular}
}
\caption{\label{tab:c_xcsqa}
   The complete experimental results on X-CSQA datasets. Avg. represents the average accuracy across all languages. 
  }
\end{table*}

\begin{table*}
\resizebox{\textwidth}{!}{
\begin{tabular}{l|l|cccccccccccccccc}
\hline
XNLI & Avg & Sw & Ur & Hi & Th & Ar & Tr & El & Vi & Zh & Ru & Bg & De & Fr & Es & En \\ \hline
LLaMAX2-XNLI & 76.5 & 66.7 & 65.6 & 70.3 & 66.5 & 73.5 & 71.8 & 76.8 & 77.5 & 78.3 & 80.4 & 81.6 & 82.2 & 83.1 & 84.1 & \textbf{89.7} \\
LLaMAX2-XNLI-SFT & 77.4 & 68.3 & 68.3 & 72.1 & 66.7 & 71.7 & 73.2 & 74.3 & 78.5 & 80.3 & 81.9 & 82.7 & 83.7 & \textbf{84.7} & \textbf{85.1} & 89.3 \\
LangBridge-SFT & 76.0 & 72.2 & \textbf{72.2} & 73.4 & \textbf{74.3} & 75.0 & 74.5 & 77.2 & 75.4 & 75.9 & 77.1 & 78.2 & 77.4 & 78.0 & 78.5 & 80.8 \\
MindMerger & 79.2 & 72.7 & 71.5 & \textbf{74.8} & 73.3 & 77.0 & \textbf{76.3} & 78.8 & 80.4 & 80.5 & 80.8 & 82.4 & 83.0 & 84.2 & 84.5 & 88.5 \\
\mname & \textbf{79.7} & \textbf{73.0} & 71.0 & 74.7 & 74.1 & \textbf{77.6} & 76.0 & \textbf{79.6} & \textbf{80.8} & \textbf{80.8} & \textbf{81.8} & \textbf{83.4} & \textbf{83.9} & \textbf{84.7} & 84.8 & 88.9 \\ \hline
\end{tabular}
}
\caption{\label{tab:c-xnli}
    The complete experimental results on XNLI datasets. Avg. represents the average accuracy across all languages. 
  }
\end{table*}

\begin{table*}
\resizebox{\textwidth}{!}{
\begin{tabular}{l|l|ccc|cccccccccc}
\hline
MGSM & parm(M) & Avg. & Lrl. & Hrl. & Bn & Th & Sw & Ja & Zh & De & Fr & Ru & Es & En \\ \hline
m-GPT & 1418 & 48.5 & 30.8 & 56.1 & 32.0 & 25.6 & 34.8 & 44.8 & 50.0 & 60.0 & 57.2 & 58.0 & 58.0 & 64.4 \\
XGLM & 1733 & 51.1 & 42.4 & 54.8 & 42.4 & 41.2 & 43.6 & 46.0 & 48.0 & 55.6 & 58.0 & 55.2 & 56.8 & 64.0 \\ \hline
nllb-3.3B & 1733 & 55.3 & 50.8 & 57.2 & 50.0 & 47.6 & 54.8 & 51.2 & 53.2 & 56.8 & 60.4 & 56.4 & 58.8 & 63.6 \\
mT5-xl & 1670 & 59.0 & 56.4 & 60.2 & 51.6 & 59.2 & 58.4 & 52.0 & 56.0 & 62.0 & 61.6 & 61.6 & 61.6 & 66.4 \\ \hline
\end{tabular}
}
\caption{\label{tab:c-encoder}
    \mname using different multilingual models as encoder and MetaMath as LLM on the MGSM dataset. Parm(M) represents the number of parameters used in the external model. Lrl., Hrl., and Avg. represent the average accuracy across low-resource languages, high-resource languages, and all languages, respectively.
  }
\end{table*}

\begin{table*}
\resizebox{\textwidth}{!}{
\begin{tabular}{l|ccc|cccccccccc}
\hline
MGSM & Avg. & Lrl. & Hrl. & Bn & Th & Sw & Ja & Zh & De & Fr & Ru & Es & En \\ \hline
w/o Adapter & 44.1 & 15.9 & 56.2 & 15.2 & 20.8 & 11.6 & 47.2 & 49.2 & 56.4 & 57.6 & 56.8 & 60.4 & 65.6 \\
w/o LLM Input & 56.8 & 55.5 & 57.3 & 50.4 & 59.6 & 56.4 & 49.6 & 54.0 & 56.4 & 60.8 & 60.0 & 58.0 & 62.4 \\
w/o Layer-Wise Aligner & 56.9 & 53.1 & 58.6 & 51.6 & 52.8 & 54.8 & 52.4 & 51.2 & 58.4 & 58.0 & 58.8 & 64.0 & 67.2 \\ \hline
w/o Translation Stage & 52.0 & 38.9 & 57.5 & 34.8 & 39.2 & 42.8 & 48.8 & 52.8 & 56.8 & 62.0 & 57.6 & 61.2 & 63.6 \\
w/o Task Stage & 38.8 & 24.7 & 44.9 & 22.8 & 22.8 & 28.4 & 30.8 & 32.0 & 51.6 & 46.4 & 42.8 & 49.6 & 60.8 \\ \hline
MetaMath & 37.9 & 5.9 & 51.6 & 6.4 & 6.4 & 4.8 & 34.8 & 39.2 & 56.4 & 55.6 & 51.6 & 55.2 & 68.4 \\
\mname & 59.0 & 56.4 & 60.2 & 51.6 & 59.2 & 58.4 & 52.0 & 56.0 & 62.0 & 61.6 & 61.6 & 61.6 & 66.4 \\ \hline
\end{tabular}
}
\caption{\label{tab:c-ablation}
    Ablation experiments of \mname on the MGSM dataset. Lrl., Hrl., and Avg. represent the average accuracy across low-resource languages, high-resource languages, and all languages, respectively.
  }
\end{table*}

\begin{table*}
\resizebox{\textwidth}{!}{
\begin{tabular}{cccccccccccccc}
\hline
\multicolumn{1}{l|}{MGSM} & Avg. & Lrl. & \multicolumn{1}{l|}{Hrl.} & Bn & Th & Sw & Ja & Zh & De & Fr & Ru & Es & En \\ \hline
\multicolumn{1}{l|}{LangBridge} & 49.1 & 44.4 & \multicolumn{1}{l|}{51.1} & 38.0 & 49.6 & 45.6 & 32.8 & 43.6 & 52.4 & 54.8 & 52.8 & 59.6 & 61.6 \\
\multicolumn{1}{l|}{\mname} & 38.1 & 5.3 & \multicolumn{1}{l|}{52.1} & 7.2 & 5.2 & 3.6 & 33.2 & 44.8 & 57.2 & 53.2 & 52.4 & 56.8 & 67.2 \\
\multicolumn{1}{l|}{\mname w/o LLM input} & 51.8 & 45.7 & \multicolumn{1}{l|}{54.5} & 42.0 & 47.2 & 48.0 & 39.6 & 44.4 & 59.2 & 53.2 & 58.8 & 62.4 & 63.6 \\ \hline
\end{tabular}
}
\caption{\label{tab:c-english}
    Experiments on MGSM using English-only task data. Lrl., Hrl., and Avg. represent the average accuracy across low-resource languages, high-resource languages, and all languages, respectively.
  }
\end{table*}

\end{CJK*}
\end{document}